\documentclass{article}

\PassOptionsToPackage{numbers, sort&compress}{natbib}

     \usepackage[preprint]{neurips_2019}
     \setcitestyle{numbers}

\usepackage[utf8]{inputenc} 
\usepackage[T1]{fontenc}    
\usepackage{hyperref}       
\usepackage{url}            
\usepackage{booktabs}       
\usepackage{amsfonts}       
\usepackage{nicefrac}       
\usepackage{microtype}      
\usepackage{graphicx}

\usepackage{algorithm}
 \usepackage{algorithmic}

 \usepackage{amsmath}

\DeclareMathOperator*{\argmax}{\arg\!\max}
 \DeclareMathOperator{\sign}{sgn}
 
\usepackage{amssymb}
\usepackage{amsthm,thm-restate} 
\usepackage{bm}
\usepackage{bbm} 
\usepackage{xcolor}
 \usepackage{mathtools}
 
 \usepackage{algorithm}
 \usepackage{algorithmic}
 
\theoremstyle{exampstyle}

\usepackage{hyperref}

\title{Iterative Policy-Space Expansion in \\
Reinforcement Learning}

\author{%
  Jan M.~Lichtenberg \\
  Department of Computer Science \\
  University of Bath\\
  Bath, United Kingdom \\
  \texttt{j.m.lichtenberg@bath.ac.uk} \\
  \And
  \"{O}zg\"{u}r \c{S}im\c{s}ek \\
  Department of Computer Science \\
  University of Bath\\
  Bath, United Kingdom \\
  \texttt{o.simsek@bath.ac.uk} \\
}

\begin{document}

\maketitle

\begin{abstract}
Humans and animals solve a difficult problem much more easily when they are presented with a sequence of problems that starts simple and slowly increases in difficulty. We explore this idea in the context of reinforcement learning. Rather than providing the agent with an externally provided curriculum of progressively more difficult tasks, the agent solves a single task utilizing a decreasingly constrained policy space. The algorithm we propose first learns to categorize features into positive and negative before gradually learning a more refined policy. Experimental results in Tetris demonstrate superior learning rate of our approach when compared to existing algorithms.
\end{abstract}

\section{Introduction}
In 1772, Benjamin Franklin received a plea for advice on a difficult career decision from his friend and fellow scientist Joseph Priestley~\citep{franklin_writings_1987}. In his reply, he described what later became known as \textit{Franklin's rule}~\citep{gigerenzer_simple_1999}:
\begin{quote}
``My way is to divide half a sheet of paper by a line into two columns, writing over the one \textit{Pro}, and over the other \textit{Con}.  [\dots] I put down under the different heads short hints of the different motives that at different times occur to me for or against the measure. When I have thus got them all together in one view, I endeavor to estimate their respective weights [\dots] If I judge some two reasons con equal to some three reasons pro, I strike out the five; and thus proceeding I find at length where the balance lies [and] come to a determination accordingly.'' (p. 878 in~\citep{franklin_writings_1987})
\end{quote}
Almost 250 years later, pros-and-cons lists are still used extensively. What makes such a simple tool so effective? One aspect could be that the main problem of estimating the relative importance of arguments is facilitated by first solving the much simpler subproblem of deciding---for each feature individually---whether it is positively or negatively associated with the response. 

We present a reinforcement learning algorithm that is similar to Franklin's rule in that the algorithm first learns, for each feature individually, whether it is positively or negatively associated with good decision outcomes. Building on these so-called \textit{feature directions}, the algorithm then gradually learns a more refined policy.

The idea that a sequence of progressively more difficult tasks could accelerate learning has been exploited in animal training where it is called \textit{shaping}~\citep{skinner_reinforcement_1958, peterson_day_2004, krueger_flexible_2009, bengio2009curriculum}. Previous research has raised the question of whether learning machines could benefit from similar ideas. In robotics, learned dynamics from regions of easy solvability are reused in more difficult regions of the task environment~\citep{sanger_neural_1994}. In \textit{curriculum learning}~\citep{elman_learning_1993, bengio2009curriculum}, neural networks are trained with progressively more noisy and less relevant training data. However, finding a good curriculum is a difficult problem and solutions are often task-specific (but see~\citep{graves_automated_2017}). 

The proposed algorithm does \textit{not} require an external teacher who guides the learning agent with a carefully tailored curriculum of tasks with increasing difficulty. The task difficulty is instead regulated intrinsically along the following two dimensions. First, the agent initially learns weights na{\"i}vely (as in \textit{na{\"i}ve Bayes}), that is, without considering interdependencies among features. Eventually, weights are estimated jointly. Second, the agent learns in a decreasingly constrained policy space, which is consistent with the view of the cognitive mechanism of humans and great apes that initially has very low capacity but grows during development~\citep{krueger_flexible_2009, brown_encoding_2005, miller_integrative_2001}.

\section{Background: Classification-based reinforcement learning with rollouts.}

\textbf{Preliminaries and Notation.} We consider a discounted Markov decision process (MDP), defined by $(\mathcal{S}, \mathcal{A}, \mathcal{G}, r, \gamma)$, where $\mathcal{S}$ is a finite set of states, $\mathcal{A}$ is a finite set of actions, $\mathcal{G}(s,a): \mathcal{S} \times \mathcal{A}(s) \rightarrow \mathcal{S} \times \mathbb{R}$ is a generative model of the environment used to sample a new state $s'$ and reward $r$ for a given state-action pair $(s, a)$, 
$r: \mathcal{S} \rightarrow \mathbb{R}$ is the reward function, and $\gamma \in (0, 1)$ is the discount factor. $\mathcal{A}(s)$ denotes the set of actions available in state $s$. The goal of reinforcement learning is to find a policy $\pi: \mathcal{S} \rightarrow \mathcal{A}$ that maximizes the expected cumulative reward of the agent.  This article is concerned with learning linear policies of the form $\pi(s) =  \underset{a \in \mathcal{A}(s)}{\argmax} \ \bm{\beta}^T \bm{\phi}(s, a)$,
where $\bm{\phi}(s,a) \in \mathbb{R}^p$ denote feature values that correspond to selecting action $a$ in state $s$ and $\bm{\beta} \in \mathbb{R}^p$ denotes the vector of feature weights to be estimated. Note that only relative differences between weights matter because the policy remains unchanged when all policy weights are multiplied with the same positive scalar. 

Policy iteration~\citep{howard_dynamic_1960,sutton_reinforcement_2018} is a classic dynamic programming method that generates a sequence of monotonically improving policies $\pi_0, \dots, \pi_k$, by alternating between two steps: estimating the value function of the current policy (\textit{policy evaluation}) and computing a new improved policy based on the current value function (\textit{policy improvement}). Large MDPs require the use of function approximation of policy and value function. The resulting algorithm is called approximate policy iteration (API).

\textbf{Classification-based reinforcement learning with rollouts.} Our work builds on a range of approximate policy iteration 
algorithms that cast the policy-improvement step as a classification problem~\citep{lagoudakis_reinforcement_2003,fern_approximate_2004,li2007focus,lazaric2016analysis,scherrer_approximate_2015,lichtenberg2019regularization}. 
A training instance of the classification data set is generated as follows, using \textit{rollouts}.
For a given state $s$, the value of an available action $a$ is approximated by the cumulative sum of rewards obtained in a finite-length forward simulation of the environment, choosing action $a$ in state $s$ and following the current policy thereafter.
The action that yields the highest cumulative reward (averaged across multiple rollouts for each action) becomes the class label for state $s$. A policy is trained to assign the correct class label to each state in the training set.
The new policy is then used in the subsequent rollout. In each iteration, the rollout starting states are sampled from a large, pre-computed \textit{rollout set}, which sometimes is generated by an existing expert policy~\citep{scherrer_approximate_2015}.

Initial versions of classification-based API algorithms worked well in problems such as learning to balance a bicycle~\citep{lagoudakis_reinforcement_2003} or in planning domains~\citep{fern_approximate_2004}.  
\textit{Classification-based modified policy iteration} (CBMPI,~\citep{gabillon_approximate_2013,scherrer_approximate_2015}) was the first RL algorithm to achieve good results in the challenging domain of Tetris. CBMPI approximates both a policy and a state-value function. The state-value function improves the accuracy of rollout estimates but its estimation requires large training sets. 

\textit{M-learning}~\citep{lichtenberg2019regularization} does not use a pre-computed rollout set. Instead, rollouts are computed exclusively for the current state of the environment, meaning that only one training instance is added to the classification data set in every iteration.
In earlier work~\citep{lichtenberg2019regularization}, M-learning was given prior knowledge about feature directions, which helped to compensate for the smaller training data available to the classifier. The direction of a feature is the sign of the corresponding weight. This prior knowledge was used in two ways. 
First, feature directions were used to identify and filter out dominated~\citep{simsek_why_2016} actions during rollouts.
Second, multinomial logistic regression, which plays a central role in M-learning, was regularized using \textit{shrinkage toward equal weights} (STEW)~\citep{lichtenberg2019regularization} regularization. The STEW penalty shrinks weights toward each other, resulting in an equal-weighting model~\citep{dawes_linear_1974, einhorn_unit_1975, gigerenzer_simple_1999, demiguel_optimal_2009} in the limit of infinite regularization. Previous research has shown the surprising effectiveness of equal-weighting models when feature directions were known in advance~\citep{wainer_estimating_1976, gigerenzer_simple_1999, demiguel_optimal_2009, lichtenberg_simple_2017}. Using prior knowledge of feature directions, M-learning was shown to learn strong Tetris policies, while using considerably fewer training samples than CBMPI~\citep{lichtenberg2019regularization}. 

When knowledge about feature directions is not available, the signs of feature weights are usually estimated implicitly as part of the general estimation procedure. We report results in Section \ref{sec:exp} that show that the absence of this prior knowledge leads to considerably slower learning performance in M-learning. The magnitude of this effect surprised us, given that the isolated estimation of feature directions in supervised learning is a relatively easy task, as supported by experimental~\citep{katsikopoulos_robust_2010, dana_superiority_2004} and theoretical~\citep{simsek_learning_2015} evidence. Motivated by this discrepancy, we decouple the estimation of directions from the remaining estimation procedure and explore a hierarchical approach that  learns feature directions first, and then builds on these directions to learn weight magnitudes. This requires an algorithm to learn feature directions.
 
\section{Learning feature directions (LFD) in reinforcement learning} \label{sec:lfd}
We present a reinforcement learning algorithm, named LFD, that learns feature directions. The feature directions ${d_i \in \{-1, 0, 1\}, \ i = 1, \dots p}$ are initialized to zero; they are said to be \textit{undecided}. The agent navigates the environment using a rollout mechanism to select actions and keeps track of how often each feature is associated positively and negatively with selected actions. A feature is assigned a direction when the difference between positive and negative associations is deemed to be significant. The algorithm terminates when all feature directions have been decided. Pseudo-code for LFD is provided in Algorithm \ref{alg:lfd} in the Appendix. A more detailed description of the algorithm follows next. 

Let $\tilde{a}$ denote the action chosen by the rollout procedure and let $\bm{\phi}(s, a_1), ..., \bm{\phi}(s, a_{|\mathcal{A}(s)|})$ denote the feature values of all actions available in state $s$. Furthermore, let $\sign$ denote the mathematical sign function: $\sign(x)$ is $1$ if $x > 0$, $0$ if $x = 0$, and $-1$ if $x < 0$. A training instance $\Delta_i$ for feature $\phi_i$ compares the feature values of the selected action $\tilde{a}$ to the feature values of all other actions. A training instance can be positive or negative and is defined as $\Delta_i = \sign\Big(\sum_{a \neq \tilde{a}} \sign\big(\phi_i(s, \tilde{a}) - \phi_i(s, a)\big)\Big).$
For example, a positive training instance means that feature $\phi_i$ was larger for the chosen action $\tilde{a}$ than for other actions more often than it was smaller. 
	
Let $n_i^+$ denote the number of positive training instances and let $n_i^-$ denote the number of negative training instances.
A feature is assigned a direction only after the difference between $n_i^+$ and $n_i^-$ is found to be statistically significant. We use a two-sided exact binomial test (for example,~\citep{howell_statistical_2009}) with null hypothesis that feature $\phi_i$ has no direction, that is, $H_0\colon  n_i^+ / (n_i^+ + n_i^-) = 0.5$. If the resulting p-value is smaller than some pre-defined threshold $\alpha$, the feature is assigned the direction $d_i = \sign(n_i^+ - n_i^-)$. 

The rollout policy utilizes features for which a direction has already been determined, while ignoring features with undecided directions. It is defined as 
$\pi_r(s) =  \underset{a \in \mathcal{A}(s)}{\argmax} \ \bm{d}^T \bm{\phi}(s, a)$, where $\bm{d} = (d_1, \dots, d_p)$ is the vector of current directions.
Ties are broken at random. 

\section{Iterative policy-space expansion (IPSE)}

By combining LFD and M-learning with STEW regularization, we create a reinforcement learning algorithm that decouples the estimation of weight signs from the estimation of weight magnitudes. LFD is employed first, until all feature directions are learned. The algorithm then switches to M-learning, treating the learned directions as useful prior knowledge.

Under conditions defined further below, the combined algorithm learns in a monotonically expanding policy space. We call this algorithm \textit{iterative policy-space expansion}, or IPSE. Note that building blocks other than LFD or M-learning could be used to create algorithms that learn in monotonically expanding policy-spaces (discussed in Section \ref{sec:dis}). To simplify notation, we will use the acronym IPSE to refer to the version that uses LFD and M-learning with STEW penalty in the remainder of this article. 

\textbf{The policy space of IPSE.} We derive necessary conditions on the regularization strength of the STEW penalty (controlled by the parameter $\lambda$) that ensure that IPSE learns in monotonically expanding policy spaces. The policy space at any given iteration is characterized by the values that the policy weight vector $\bm{\beta}$ can attain. Initially, IPSE learns directions using the LFD algorithm; the policy space is therefore constrained to be $\Pi_{\text{LFD}} = \{-1, 1\}^p$. During the M-learning phase, the policy space is a function of the regularization strength $\lambda >0$. STEW-regularized multinomial logistic regression can be reformulated as a constrained optimization problem (similar to, for example,~\citep{van_wieringen_lecture_2015}) to see that the policy space is given by ${\Pi_{\lambda} = \{\bm{\beta} \in \mathbb{R}^p | \sum_{i=1}^p (\beta_i - d_i)^2 \leq c(\lambda)\}}$, where $d_i$ are the directions estimated by LFD in the first phase of the algorithm, and $c(\lambda)\colon \mathbb{R}_0^+ \rightarrow \mathbb{R}_0^+$ is a decreasing function of $\lambda$, for which the following holds true: $c(\lambda) \rightarrow 0$ for  $\lambda \rightarrow \infty$; and $c(\lambda) \rightarrow 0$ for $\lambda \rightarrow \infty$. The policy space therefore is a hypersphere around the equal-weights solution that was found by the LFD algorithm. The size of that hypersphere is a decreasing function of the regularization strength $\lambda$. Let $\{\lambda_k\}_{k=1}^\infty$ denote a sequence of decreasing regularization strengths. It then follows that $\Pi_{\text{LFD}} \subset \Pi_{\lambda_k} \subset \Pi_{\lambda_{k+1}} \subset \mathbb{R}^p,$ or in other words, the policy space is monotonically expanding.

\textbf{Choice of $\lambda$.} In practice (for example, in earlier work on M-learning~\citep{lichtenberg2019regularization}), the regularization strength is often chosen using cross validation. Here, we use a pre-defined schedule of decreasing regularization strengths in order to ensure a monotonically expanding policy space. We aim to find a schedule that satisfies the following two properties. 
First, the regularization strength should initially be high enough to ensure a smooth transition from LFD to M-learning. Second, the regularization strength should decrease rapidly enough so that the policy space is not overly constrained for too long. Both these properties are satisfied in the following example. 

\begin{figure}
 \centering
   \includegraphics[width=0.8\linewidth]{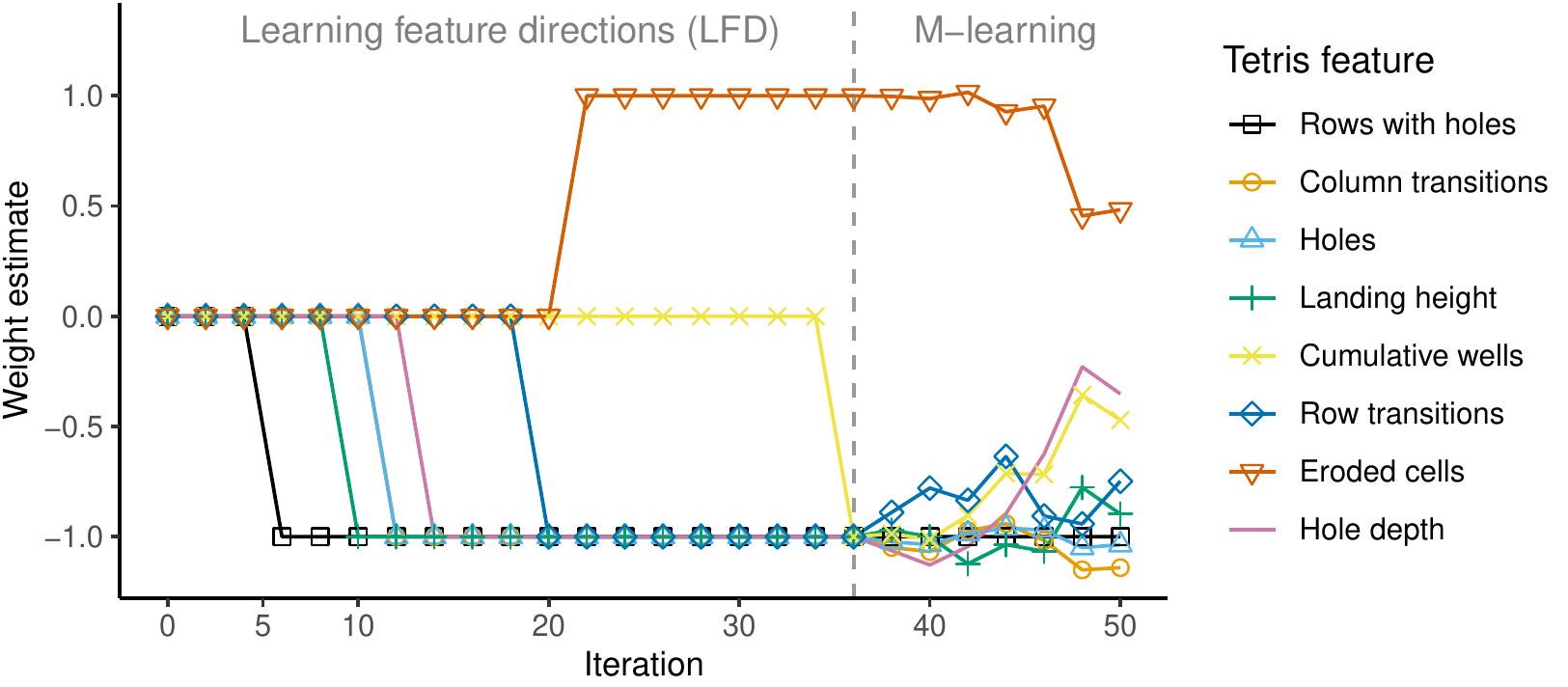} 
\caption{Policy weight trajectories of the IPSE algorithm in Tetris. The dashed vertical line signifies the transition from the LFD algorithm to M-learning with STEW penalty.
Weights were rescaled such that the weight \textit{rows with holes} always has an absolute value of 1.
}
\label{fig:weights}
\end{figure}

\textbf{Example weight trajectories.} Figure \ref{fig:weights} shows policy weight trajectories of the IPSE algorithm obtained while learning to play Tetris (see Section \ref{sec:exp} for a detailed description of the experimental setup). IPSE used  $\lambda_k = 5/k$ in the $k$-th iteration of M-learning. In the iterations directly following the transition to M-learning (iteration 36 in this particular example), the estimated weights remained relatively close to the equal-weighting solution. Policy weight estimates then increasingly deviated from the equal-weights solution as the policy space expanded.

\section{Experiments} \label{sec:exp}
We next present results from our experiments in Tetris. We used an experimental setup similar to the one that  was used to demonstrate the fast learning rate of M-learning~\citep{lichtenberg2019regularization}, with the important difference that in our experiments, feature directions (that is, weight signs) were \textit{not} given in advance. 
Our primary objective is to examine whether IPSE benefits from learning weight signs and magnitudes sequentially rather than jointly, as is done by competing algorithms. 

\textbf{Tetris.} Tetris can be formulated as a MDP, where the state consists of the board configuration and the identity of the falling tetrimino. Available actions are the possible placements of the tetrimino on the board. 
A reward of 1 is received for each cleared line. The game ends when a state allows no legal placement. 
We used a board size of $10\times10$ in all experiments. Eight features were used to describe a state-action pair: \textit{landing height}, \textit{number of eroded piece cells}, \textit{row transitions}, \textit{column transitions}, \textit{number of holes}, \textit{number of board wells}, \textit{hole depth}, and \textit{number of rows with holes}. These features are from earlier work by \citet{thiery_building_2009}, who describe them in detail. 

\textbf{Algorithms.} 
We compared IPSE to the LFD algorithm, CBMPI, and four versions of M-learning. The M-learning versions  differed in regularization behavior and prior knowledge available. One version did not make use of regularization at all. The other three versions used STEW regularization. Among the three regularizing versions, one used cross-validation to estimate $\lambda$ (as in~\citep{lichtenberg2019regularization}), while the other two used a schedule as described in the previous section. Among the latter two versions, one was given knowledge about feature directions obtained from the weights of the BCTS policy~\citep{thiery_building_2009}. 

All M-learning versions, IPSE, and LFD used the same rollout parameters. These algorithms computed $M=10$ rollouts of length $T=10$ for each action (compare to Algorithm \ref{alg:rollout} in the Appendix). Given that the number of actions is always smaller than $34$, the maximum number of calls to the generative model of Tetris for one iteration of the algorithm was at most $34TM = 3400$. We used a per-iteration budget of 170,000 calls for CBMPI. The Appendix contains further implementation details.

\begin{figure}
 \centering
   \includegraphics[width=0.9\linewidth]{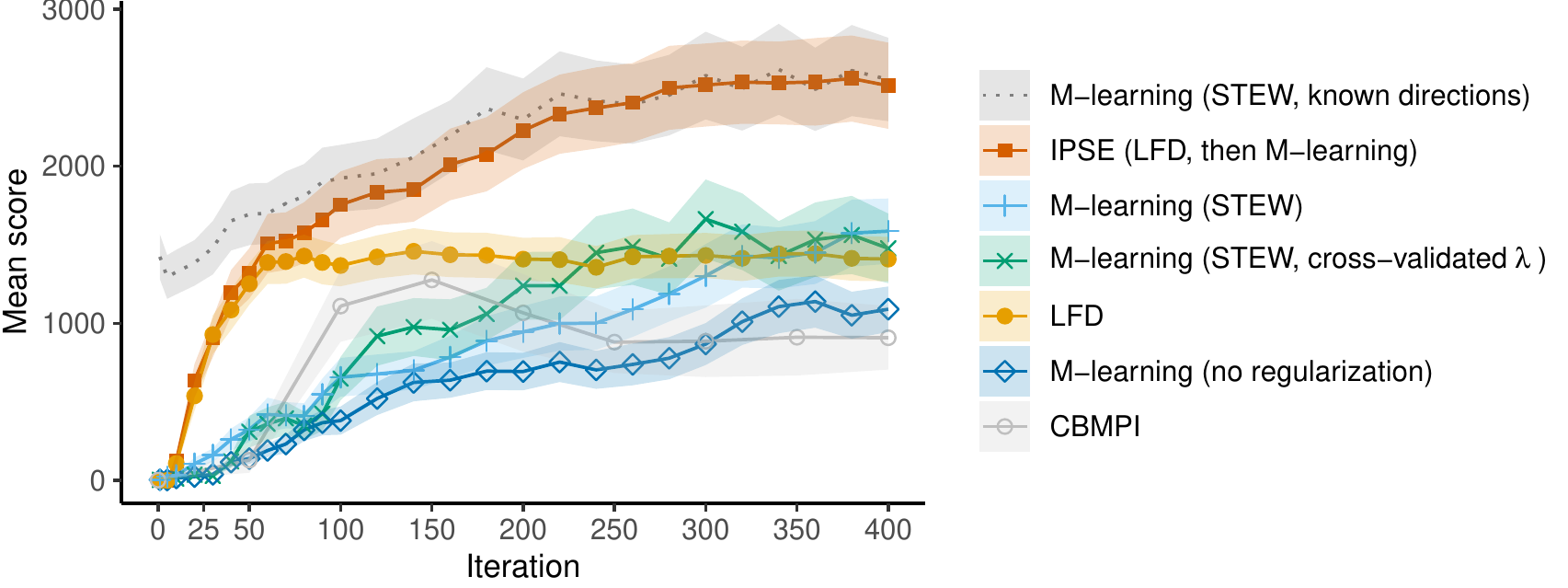} %
\caption{Quality of the policy learned as a function of the iterations of the algorithm. Each learning curve shows means across 100 replications of the algorithm. Quality of the policy is measured by the mean score obtained by the policy in 30 Tetris games. 
}
\label{fig:rl}
\end{figure} 

\textbf{Results.} 
Learning curves are shown in Figure \ref{fig:rl}. M-learning with given feature directions represents an upper baseline. This is the dotted line in the figure. 
Among algorithms that were not given prior knowledge about feature directions, IPSE showed the highest learning rate and learned the best policies overall. Furthermore, it rapidly approached the ceiling performance obtained with known feature directions. All other algorithms learned more slowly. At 400 iterations, there was a large performance gap between IPSE and all other algorithms. 

The hypothesis that IPSE benefits from learning directions independently was supported by the strong performance of the standalone LFD algorithm at the beginning of the learning curve. This indicates that IPSE could fruitfully use the na{\"i}ve direction estimates as a stable basis for later learning.

\section{Discussion and future work}  \label{sec:dis}
Our experimental results show that reinforcement learning algorithms can benefit from learning in a policy space that initially is strongly constrained but expands during the learning process. Similar to how people structure their thoughts by categorizing arguments into \textit{pro} and \textit{contra}, learning feature directions has proven to be a useful building block for learning more complex policies. 

An interesting direction for future work is to extend the approach presented in this paper to reinforcement learning with non-linear function approximators such as neural networks. These algorithms estimate a much larger number of feature weights, which requires more data, and thus makes a hierarchical approach potentially even more promising. However, it is unclear whether the notion of feature directions is useful for certain network architectures such as convolutional neural networks. 
 
\small

\bibliography{ipse_preprint}
\bibliographystyle{unsrtnat}

\appendix
\section*{Appendix}

\section{Additional implementation details}

An overview of machine learning solutions to Tetris can be found in~\citet{algorta_game_2019}.
We used implementations of Tetris and the M-learning algorithm  from~\citet{lichtenberg2019regularization}.

\textbf{M-learning.} Multinomial logistic regression in iteration $k$ used the most recent $n(k)$ training samples, where $n(k) = min(100, \lfloor \frac{k}{2} \rfloor+2)$. 

\textbf{LFD.} We used alternative rollout policy that uses the rollout policy $\pi_r(s)$ (described in Section \ref{sec:lfd}) unless an immediate reward greater than zero for at least one action is possible, in which case the action that promises the highest reward is selected. We found this alternative rollout policy to have a positive effect on the learning rate in all of our experiments.

\textbf{CBMPI.} The CBMPI results reported by \citet{scherrer_approximate_2015} used a per-iteration budget of 8,000,000 calls to the generative model of Tetris. In comparison, the total budget (after 400 iterations) we used for the other algorithms was 1,360,000. In order to compare the algorithms meaningfully, we experimented with CBMPI using budgets in the same range as M-learning. We present results with CBMPI using a per-iteration budget of 170,000.
\section{Pseudo-code}

\begin{minipage}[t]{0.54\textwidth}
\vspace{0pt}
\begin{algorithm}[H] 
\begin{algorithmic}
\footnotesize
\caption{Learning feature directions (LFD)}
      \label{alg:lfd}
   \STATE {\bfseries Output:} 
   \STATE $\bm{d}\in \{-1, 0, 1\}^p$ \hfill \textit{// feature directions, initialized to $\bm{0}$}
   \STATE {\bfseries Input: }
   \STATE $\alpha \in (0, 1)$ \hfill \textit{// significance threshold}
   \STATE $\pi_r(s, \bm{d}): \mathcal{S} \times \{-1, 0, 1\}^p \rightarrow  \mathcal{A}$ \hfill \textit{// rollout policy using}
   \STATE  \hfill \textit{current feature directions}
\STATE
   \STATE $s \leftarrow $ state sampled from initial state distribution 
  \WHILE{not all directions are learned} 
 	 \FORALL{$a \in \mathcal{A}(s)$}
		   \STATE  $\hat{U}(s, a) \leftarrow$ \textsc{Rollout}($s, a,\pi_r(s, \bm{d})$) 
    	\ENDFOR
   	\STATE $\tilde{a} \leftarrow \underset{a \in \mathcal{A}(s)}{\argmax}\  \hat{U}(s, a)$ 
	\STATE Take action $\tilde{a}$ and observe new state $s'$
	\IF{$s'$ is not terminal}
		\FORALL{$i = 1, \dots, p$}
			\STATE $\Delta_i = \sign\Big(\sum_{a \neq \tilde{a}} \sign\big(\phi_i(s, \tilde{a}) - \phi_i(s, a)\big)\Big)$
			\STATE $n_i^+ \leftarrow n_i^+ + \max(\Delta_i, 0)$
			\STATE $n_i^- \leftarrow n_i^- - \min(\Delta_i, 0)$
			\STATE \textit{p-val} $\leftarrow$ test $H_0\colon  n_i^+ / (n_i^+ + n_i^-) = 0.5$
			\IF{\textit{p-val} < $\alpha$}
				\STATE $ d_i \leftarrow
				    	\begin{cases*}
				      		   1 & if $n_i^+ > n_i^-$ \\
      						  -1  & otherwise
					\end{cases*}$
				
			\ENDIF
		\ENDFOR
	  	\STATE $s \leftarrow s'$
        \ELSE
\STATE         \textit{// reset episode}
  		\STATE $s \leftarrow $ state sampled from initial state distribution 
	\ENDIF
 \ENDWHILE
   \end{algorithmic}
\end{algorithm}
\end{minipage} 
\begin{minipage}[t]{0.455\textwidth}
\vspace{0pt}
\begin{algorithm}[H] 
\begin{algorithmic}
\footnotesize
\caption{\textsc{Rollout}($s, a,  \pi_r$)}
\label{alg:rollout}
\STATE {\bfseries Output:} 
\STATE $\hat{U} \in \mathbb{R},$ \textit{ estimated value of taking action $a$ in $s$}
\STATE {\bfseries Input:} 
\STATE $s \in \mathcal{S} $ \hfill \textit{// rollout starting state} 
\STATE $a \in \mathcal{A}(s)$ \hfill \textit{// action to be evaluated}
\STATE $\pi_r(s): \mathcal{S} \rightarrow  \mathcal{A}$ \hfill \textit{// rollout policy}
\STATE $M \in \mathbb{N}$ \hfill \textit{// number of rollouts }
\STATE $T \in \mathbb{N} $ \hfill \textit{// rollout length}
\STATE $\gamma \in [0,1] $ \hfill \textit{// discount factor }
\STATE $\mathcal{G}(s,a): \mathcal{S} \times \mathcal{A}(s) \rightarrow \mathcal{S} \times \mathbb{R}$   \hfill  \textit{// generative}
\STATE  \hfill  \textit{model}
\STATE
\FORALL{$j = 1, \dots, M$}  
	\STATE $(s', r) \leftarrow \mathcal{G}(s, a)$ 
	\STATE $\hat{U}_j \leftarrow r$
	\STATE $s \leftarrow s'$
	\FORALL{$t = 1, \dots, T-1$}  
		\STATE $(s', r) \leftarrow \mathcal{G}(s, \pi_r(s))$ 
		\STATE $\hat{U}_j \leftarrow \hat{U}_j + \gamma^t r$
		\STATE $s \leftarrow s'$
	\ENDFOR
\ENDFOR
\STATE {\bfseries return} $\hat{U} \leftarrow \frac{1}{M}\sum_{j=1}^M \hat{U}_j$
\end{algorithmic}
\end{algorithm}
\end{minipage}

\begin{minipage}[t]{\textwidth}
\centering
\begin{minipage}[t]{0.9\textwidth}
\vspace{0pt}
\begin{algorithm}[H] 
\begin{algorithmic}
\footnotesize
\caption{Online reinforcement learning with rollouts (general form)}
      \label{alg:gen}
         \STATE {\bfseries Input: } 
         \STATE  $\Pi$ \hfill \textit{// policy space}
         \STATE  $\Omega$ \hfill \textit{// space of data sets produced by a rollout procedure}
         \STATE \textsc{Learn}: $\Omega \rightarrow \Pi$, where \hfill \textit{// learning procedure}
         \STATE $\mathcal{D} = \emptyset $  \hfill \textit{// data structure  to store choice data}

   \STATE {\bfseries Output:} 
   \STATE $\pi \in \Pi$ \hfill \textit{// policy, initialized to uniform random policy}

\STATE
   \STATE $s \leftarrow $ state sampled from initial state distribution 
  \WHILE{not all directions are learned} 
 	 \FORALL{$a \in \mathcal{A}(s)$}
		   \STATE  $\hat{U}(s, a) \leftarrow$ \textsc{Rollout}($s, a, \pi$) 
    	\ENDFOR
   	\STATE $\tilde{a} \leftarrow \underset{a \in \mathcal{A}(s)}{\argmax}\  \hat{U}(s, a)$ 
	\STATE Take action $\tilde{a}$ and observe new state $s'$
	\IF{$s'$ is not terminal}
	   \STATE $\mathcal{D} \leftarrow \mathcal{D} \cup \{\{\tilde{a}, \bm{\phi}(s, a_1), \bm{\phi}(s, a_2), ..., \bm{\phi}(s, a_{|\mathcal{A}(s)|})\}\}$ \hfill \textit{// append choice set to $\mathcal{D}$}
		\STATE $\pi \leftarrow \textsc{Learn}(\mathcal{D})$
	  	\STATE $s \leftarrow s'$
        \ELSE
  		\STATE $s \leftarrow $ state sampled from initial state distribution \hfill \textit{// reset episode}
	\ENDIF
 \ENDWHILE
   \end{algorithmic}
\end{algorithm}
\end{minipage} 
\end{minipage}

\end{document}